# Application of Monte Carlo Stochastic Optimization (MOST) to Deep Learning

Sin-ichi Inage*, Hana Hebishima

Fluid Engineering Laboratory, Department of Mechanical Engineering, Fukuoka University, Fukuoka 8140180, Japan

*Corresponding author: Prof. Shin-ichi Inage

Department of Mechanical Engineering, Fukuoka University

8-19-1, Nanakuma, Jonan-ku, Fukuoka 814-0180, Japan

Tel: +81928716631(Ext. 6317)

Fax: +81928656031

Email: s.inage.pu@adm.fukuoka-u.ac.jp

Abstract

In this paper, we apply the Monte Carlo stochastic optimization (MOST) proposed by the authors to a deep learning of XOR gate and verify its effectiveness. Deep machine learning based on neural networks is one of the most important keywords driving innovation in today's highly advanced information society. Therefore, there has been active research on large-scale, high-speed, and high-precision systems. For the purpose of efficiently searching the optimum value of the objective function, the author divides the search region of a multivariable parameter constituting the objective function into two by each parameter, numerically finds the integration of the two regions by the Monte Carlo method, compares the magnitude of the integration value, and judges that there is an optimum point in a small region. In the previous paper, we examined the problem of the benchmark in the optimization method. This method is applied to neural networks of XOR gate, and compared with the results of weight factor optimization by Adam and genetic algorithm. As a result, it was confirmed that it converged faster than the existing method.

## 1. INTRODUCTION

Recently, neural networks are widely used as a machine learning method for various problems in the artificial intelligence field. In deep learning and neural networks, it is necessary to update parameters called weights repeatedly so that they are not captured by local solutions with large errors in loss functions. The learning process of neural networks is strongly related to the optimization of the objective function, and the selection of the optimization method is a very important step in the design of the learning algorithm. As an optimization of deep learning, there is a steepest descent method. The steepest descent method can be said to be the simplest gradient method to select and update the gradient direction in which the objective function decreases most. However, since it is necessary to calculate the sum of the difference values of n objective functions, the calculation time becomes longer as the number of data increases. The stochastic gradient descent method is a typical method to solve this problem.

The gradient descent method is the most common method that uses the gradient of the objective function to progressively reduce the value of the objective function until it is finally reached. There's still a lot of research into improved algorithms today, and a lot of excellent reviews (Noor Fatima, 2020, Prabhu Teja S, Florian Mai, 2020, Parnia Bahar et al., Peter Henderson et al., 2018). In order to improve the accuracy without falling into the local solution with large error, it seems to be important to search a wide range avoiding the local solution with large error in the first half of the learning, and to stably converge toward the solution with small error in the latter half. Typical algorithms include Momentum SGD (Rumelhart, David E.; Hinton, Geoffrey E., 1986), Nesterov accelerated gradient (NAG), AdaGrad (Duchi, J., Hazan, E., & Singer, Y., 2011), RMSprop (Tijmen Tieleman; G. Hinton,2012), AdaDelta (Zeiler, M. D., 2012), and Adam (Kingma, D. P., and Ba, J. L.,2015). Table 1 compares these algorithms.

In usual SGD, the convergence is retarded by the oscillatory fluctuation in the high gradient region of the objective function, and the convergence does not progress in the low gradient region. Momentum SGD introduces the concept of inertia to solve this problem. Inertia is the same concept as so-called moving average, and the introduction of moving average is expected to suppress the variation of convergence. In the equation of momentum SGD in Table 1, the γmt -1 part of mt corresponds to inertia and suppresses fluctuation. γ is a constant smaller than 1.0.

AdaGrad changes the learning rate for each variable axis. In the case of data rarely observed in the input data of machine learning, the gradient in the axial direction of the data is close to zero, and the convergence of the weight function becomes slow. Therefore, it is a method to increase the learning rate in this axis. In AdaGrad, the learning rate is divided by the square root of the sum of the squares of the gradients in each axial direction to increase the learning rate especially for rare features (≒ zero gradient). In order to avoid zero splitting, a small value ε is added to the denominator.

Like Momentum, RMSprop is designed to reduce the vibration of SG D. RMSprop adjusts the learning rate according to the magnitude of the slope. Like AdaGrad, it uses the sum of squares of gradients. The method of changing the learning rate also introduces the same idea as AdaGrad.

In the case of RMSprop et al., there is a problem that dimensions do not match because weight coefficients are updated by gradients. Adadelta solves this problem by using the exponential moving average of the square of the difference of the parameters, and at the same time, setting of the learning rate is unnecessary. Therefore, AdaDelta does not include tuning.

Adam can be understood as a combination of momentum SGD and RMSProp. It is expected to converge very quickly, and this algorithm has been established as one of the most popular neural network learning algorithms. RMSprop and Adadelta do not consider information about past gradients. Adam takes historical gradients and exponential moving averages and uses the unbiased invariant estimator of the exponential moving average. Adam is discussed again in Chapter 3 as a benchmarking.

Many other algorithms derived from the gradient descent algorithm are proposed and studied, such as Nadam(Timothy Dozat, 2015), AdaMax (Ganbin Zhou et al. 2017), AMSGrad (Sashank J. et al.) and Eve (Hiroaki Hayashi et al., 2018) etc. On the other hand, the author proposed a binary Monte Carlo stochastic optimization (MOST) that is different from the gradient descent method (S. Inage et al., 2021). Instead of focusing on gradients, In the proposed method, a search region is divided into two regions, and an objective function is numerically integrated in each region by Monte Carlo method. By comparing integral values in each region, it is judged which region contains the optimum solution, and by repeating it, the final optimum solution is searched. This paper first explains the concept and then applies it to neural network learning of XOR gate and compares the results with those of existing optimization methods - Adam and genetic algorithms. Details are described below.

Table 1. : Comparison of existing optimization methods on Neural Network

| Optimization methods | References |
|---|---|
| [Stochastic Gradient Descent (SGD)] $$w^{t+1} = w^t - \eta \left(\frac{\partial L}{\partial w}\right)^t$$ η: Learning rate | Bottou, Léon; Bousquet, Olivier (2012). |
| [Momentum SGD] First moment: $m^t = \gamma m^{t-1} + \eta \left(\frac{\partial L}{\partial w}\right)^t$ $w^{t+1} = w^t - m^t$ Initial value: $m_0 = 0$ | Rumelhart, David E.; Hinton, Geoffrey E. (1986) |
| [Nesterov accelerated gradient(NAG)] First moment: $m^t = \gamma m^{t-1} + \eta \left(\frac{\partial L}{\partial w}\right)(w^t - m^{t-1})$ $w^{t+1} = w^t - m^t$ | Nesterov, Y. (1983) |
| [AdaGrad] Second moment: $v_i^{t+1} = v_i^t + \left(\left(\frac{\partial L}{\partial w}\right)^t\right)^2$ $w_i^{t+1} = w_i^t - \frac{\eta}{\sqrt{v_i^t + \varepsilon}} \left(\frac{\partial L}{\partial w}\right)^t$ Initial value: $v_i^0 = 0$ | Duchi, J., Hazan, E., & Singer, Y. (2011) |
| [RMSprop] Second moment: $v_i^{t+1} = \gamma v_i^t + (1-\gamma)\left(\left(\frac{\partial L}{\partial w}\right)^t\right)^2$ $w^{t+1} = w^t - \frac{\eta}{\sqrt{\hat{v}^t} + \varepsilon} \left(\frac{\partial L}{\partial w}\right)^t$ Initial value: $v_i^0 = 0$ | Tijmen Tieleman; G. Hinton (2012) |
| [AdaDelta] Second moment: $v^{t+1} = \gamma v^t + (1-\gamma)\left(\left(\frac{\partial L}{\partial w}\right)^t\right)^2$ $s^{t+1} = \gamma s^t + (1-\gamma)(\Delta w_i^t)^2$ $\Delta w_i^t = -\frac{\sqrt{s_i^t + \epsilon}}{\sqrt{v_i^t + \epsilon}}$ $w^{t+1} = w^t + \Delta w_i^t$ Initial value: $v_i^0 = s_i^0 = 0$ | Zeiler, M. D. (2012). |
| [Adam] First moment: $m^{t+1} = \beta_1 m^t + (1-\beta_1)\left(\frac{\partial L}{\partial w}\right)^t$ Second moment: $v^{t+1} = \beta_2 v^t + (1-\beta_2)\left(\left(\frac{\partial L}{\partial w}\right)^t\right)^2$ $\widehat{m} = \frac{m^{t+1}}{1-\beta_1^{t+1}}$ $\hat{v} = \frac{v^{t+1}}{1-\beta_2^{t+1}}$ $w^{t+1} = w^t - \eta \frac{\widehat{m}^t}{\sqrt{\hat{v}^t} + \varepsilon}$ Initial value: $v_i^0 = m_i^0 = 0$ | Diederik Kingma, Jimmy Ba.(2015) |

## 2. Brief summary of Monte Carlo Optimization

### 2.1 Fundamental concept

The fundamental concept of the MOST is outlined below. Consider the following function of the univariate x.

$$f(x) = 2x^3 - \frac{5}{2}x^2 + \frac{1}{2}x + \frac{1}{2} \quad (1)$$

Assuming that the search area is x = [0,1], f(x) has a minimum value at x = 0.72, as shown in FIG. 1. The MOST first divides the search area [0, 1] into [0, 0.5] and [0.5, 1] at its midpoint. In MOST, it is used that the value integrated in each region becomes smaller on the side where the minimum value is included. The integration is performed by the Monte Carlo method using uniform random numbers defined in the domain. The values of the numerical integration of the objective function in each region can be obtained by:

$$\text{Numerical integration in region [1]:} \quad = \frac{1}{2}\sum_1^M f(x_{[1]i}) \quad (2)$$

$$\text{Numerical integration in region [2]:} \quad = \frac{1}{2}\sum_1^M f(x_{[2]i}) \quad (3)$$

In FIG. 1 a), the symbol "×" on the x-axis indicates the position of the random number. On this case, there are 10 random numbers in each region. The integral decreases on the side of the region [2] containing the minimum value. Next, a region having a small integral value is set as a new search region and divided into two again to be [1] and [2]. Compared with the initial partition width of 0.5, the new partition width is reduced by half to 0.25. Similarly, uniform random numbers are defined in each region and integrated by the Monte Carlo method. As shown in FIG. 1 b), the integral value becomes smaller in region [1]. If this area is further divided into two parts and the same calculation is repeated, the division width becomes smaller sequentially. In the present case, if the calculation is repeated N times, the division width is $1/2^N$. If N is 20, it is $1\times^{-6}$ or less. In this case, the search area converges to 0.72 within an error of $1\times10^{-6}$. This is the fundamental concept of MOST. The main features are as follows.

1)It is a stochastic method using random numbers, but optimization is a deterministic number without stochastic fluctuation.

2) The solution always converges. In particular, more than 20 MOST iterations converge to $1/10^6$ of the initial search width.

3) It does not include hyperparameters that need to be tuned for optimization. There is also no dimensional mismatch problem to be addressed in the gradient descent method.

4) There is no inherent dimensional mismatch problem in the gradient descent method.

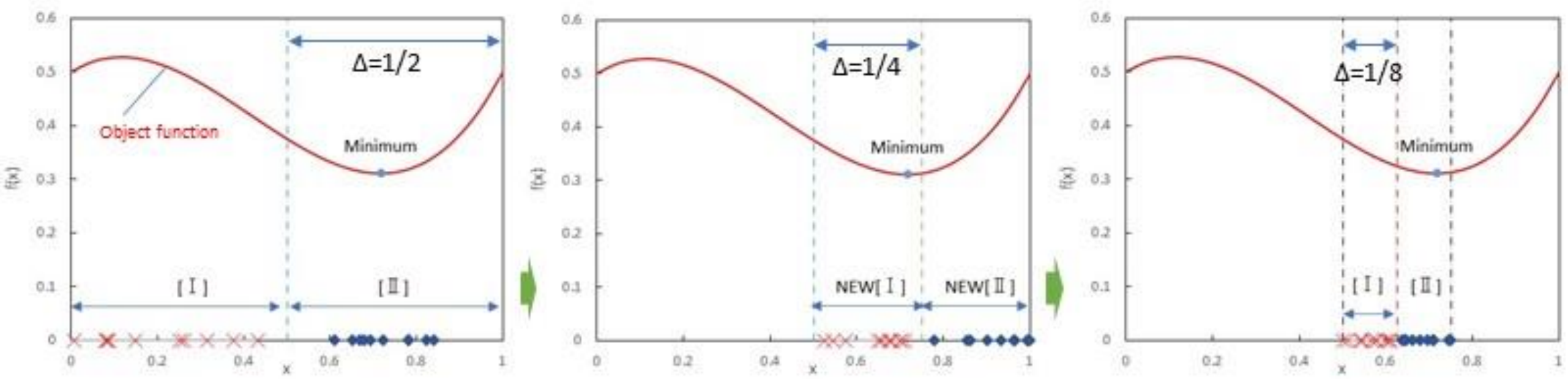

STEP-1 STEP-2 STEP-3

FIG.1 Fundamental concept of proposal algorithm

### 2.2 Extension to a multivariable objective function

The concept of MOST can be easily generalized, and even in the case of the objective function f ($x_1, x_2, \ldots, x_n$) of the multivariable $x_1, x_2, \ldots, x_n$, in the Monte Carlo method, the integral value is given by the following equation when the domain is $x_1, x_2, \ldots, x_n = [0,1]$.

$$\textit{Numerical integration} = \frac{1}{M}\sum_{j=1}^{M} f\left(x_{1j}, x_{2j}, \cdots, x_{Nj}\right) \tag{4}$$

It is well known that as the number of parameters to be integrated increases, only the Monte Carlo method is available for numerical integration. Next, the search regions of the multivariable $x_1$, $x_2$,..., $x_n$ are divided into two regions at the same time, and the above concept is applied. In this case, the number of integral calculations and their comparison is $2^n$ times per a step. For large optimization calculations, $2^n$ calculations are impractical. To solve this problem, consider dividing each variable into two sequentially as follows. In the multivariable $x_1$, $x_2$,..., $x_n$, the search region of only $x_1$ is divided into two, a uniform random number satisfying each region is considered, and the remaining $x_2$,..., $x_n$ are integrated by Monte Carlo method using a uniform random number satisfying the initial search whole region not divided. In the comparison, we first evaluate which region contains the minimum value of the objective function for the variable $x_1$. Next, a search region of $x_2$ is divided into two, and a uniform random number satisfying each region is considered. Using a random number $x_1$ that satisfies the new search region already obtained and a uniform random number $x_3$,... $x_n$ that satisfies the entire initial search region that is not divided, the integration is performed by the Monte Carlo method. In this operation, the region in which the objective function becomes minimum can be specified even for $x_2$. Repeat to $x_3$, $x_4$,..., $x_n$. This sequential bisection of $x_1$, $x_2$,..., $x_n$ is defined as one calculation of a new MOST. If the searching region of each variable is reduced by half in this flow, the number of calculations required is only $2 \times n$, and the calculation load can be significantly reduced as compared with $2^n$ described above. FIG. 2 shows a two-variable case, and FIG. 3 shows a two-division method and a calculation direction of MOST in a three-variable case.

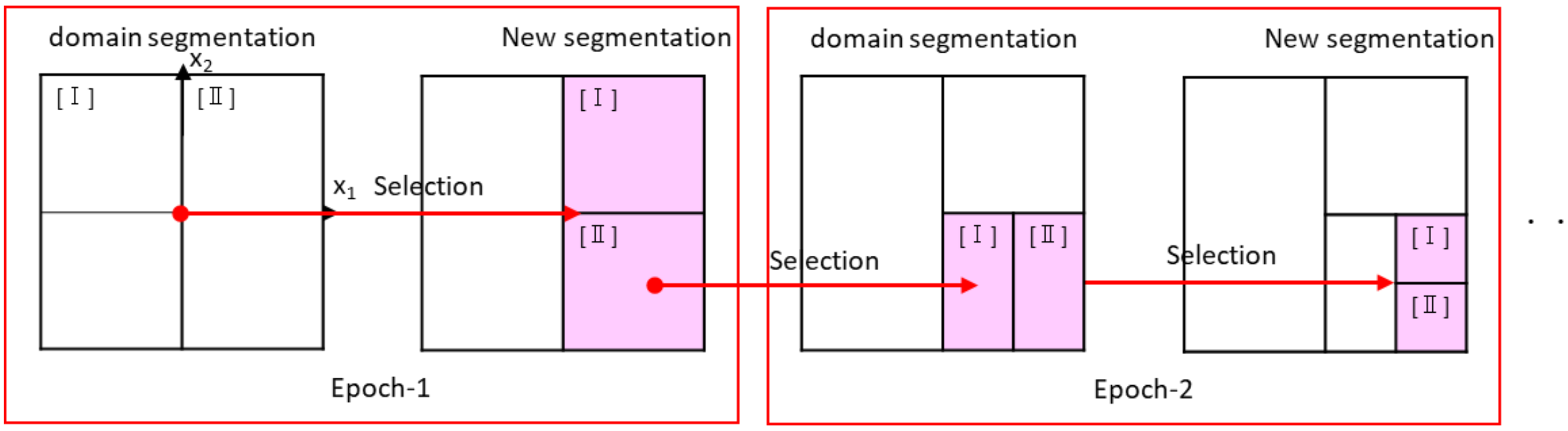

a) Modified segmentation

FIG. 2 Modification of optimization using Monte Carlo method

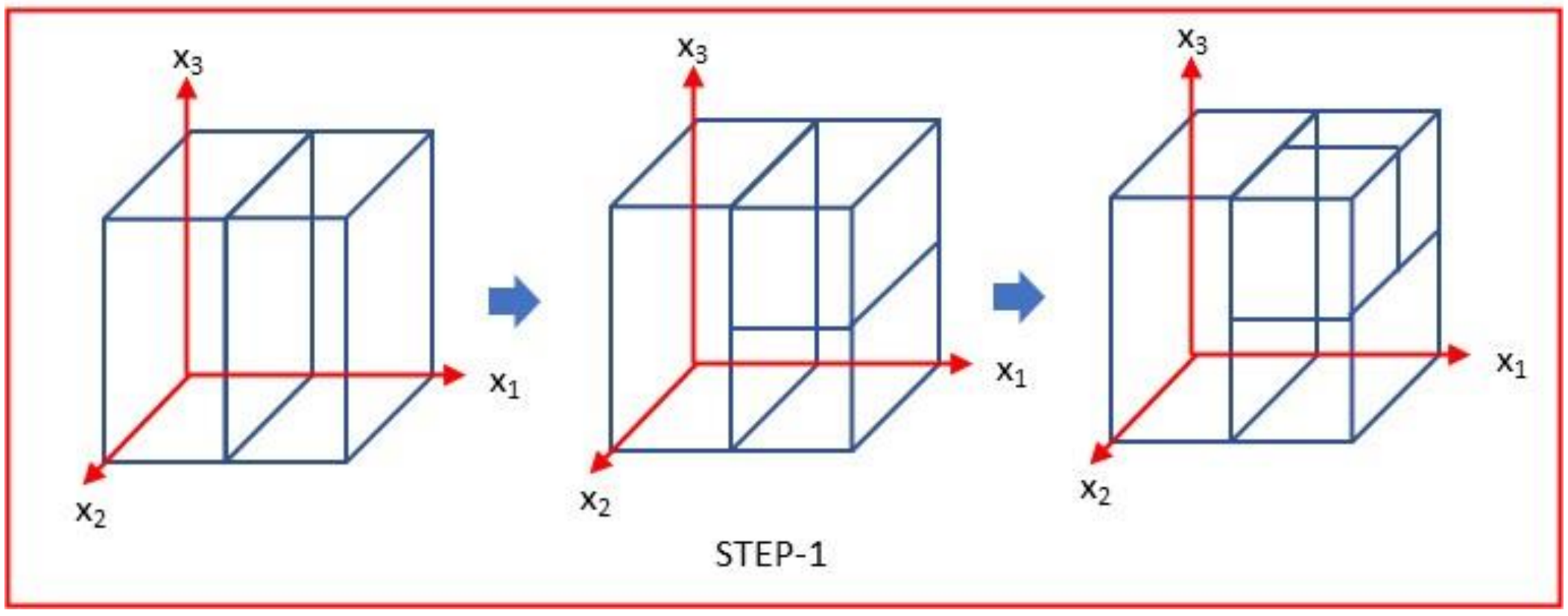

FIG. 3 Modification of optimization using Monte Carlo method for three variables

At the end of this section, the flow of the MOST algorithm is summarized below. We consider an objective function consisting of the multivariable parameters $x_1$, $x_2$, $x_3$, ・・・, and $x_n$.

1) The search ranges of the parameters $x_1$, $x_2$, $x_3$, ・・・ and $x_n$ are given.
2) First, the search range $x_1$ is divided into two regions [1] and [2]. The remaining parameters consider the entire search range.
3) For $x_1$, the uniform random numbers $x_{1[1]}$ and $x_{1[2]}$ defined in the ranges [1] and [2] above, and for the other $x_2$, $x_3$, ・・・, and $x_n$, the uniform random numbers defined in the entire search range are generated.
4) The integral values of the objective function in the ranges [1] and [2] are obtained by the Monte Carlo method as follows and compared.

   Integral value of [1] = $\sum_{i=1}^{N} f\left(x_{1[1]i}, x_{2i}, x_{3i,\dots,} x_{ni}\right)$ (5)

   Integral value of [2] = $\sum_{i=1}^{N} f\left(x_{1[2]i}, x_{2i}, x_{3i,\dots,} x_{ni}\right)$ (6)

   where i :number of a random number, N:number of Monte Carlo calculations. To calculate the integral correctly, Eqs. (5) and (6) should be multiplied by a constant determined from the maximum and minimum values of the search region. However, in MOST, since the magnitude of the integral value is important, the effect of the constant can be ignored.
5) A region with a small integral value is selected as a new search region of $x_1$.
6) Next, the search range of $x_2$ is divided into two regions [1] and [2]. $x_1$ generates a uniform random number defined by the new search area obtained in (5), and generates a uniform random number defined by the whole search area except $x_1$ and $x_2$.
7) The integral value of the objective function in the range of the search regions [1] and [2] of $x_2$ is obtained and compared by the Monte Carlo method.
8) A region with a small integral value is selected as a new search region of $x_2$. This determines a new search area of $x_1$ and $x_2$.
9) Hereinafter, process 6) to 8) above are repeated even if $x_3$ or less to find a new search area $x_3$, ・・・, and $x_n$. When a new search area is found between $x_1$ and $x_n$, the sequence of calculations is called an "Epoch".
10) When the first Epoch is completed, it is half of the first search area, and the new search area is used to calculate the second and subsequent epoch. When the search area becomes sufficiently small, for example, $<1\times10^{-6}$, the optimization calculation is completed.

### 2.2 Verification of the MOST

In this section, we applied MOST to the following Schwefel function and Sphere function as benchmark problems (Refs 17), 18)) in optimization. The number of variables to be optimized was 5.

Schwefel function: $f(x_1,\cdots,x_5) = -\sum_{i=1}^{5} x_i sin\left(\sqrt{|x_i|}\right)$ $(-500<x_1,\cdots,x_5<500)$ (7)

Sphere function: $f(x_1,\cdots,x_5) = -\sum_{i=1}^{5} x_i^2$ $(-1<x_1,\cdots,x_5<1)$ (8)

As described above, the domain of the variable was divided into two, and the Monte Carlo method was applied with a combination of 2000 random numbers in each domain. Table 2 and Table 3 show the comparison between the theoretical value and the evaluation result by MOST of $x_1$ to $x_5$ values giving the minimum point in Schwefel function and Sphere function, respectively. The number of MOST iterations is 20. In the case of Schwefel function, the error is 0.23% at the maximum and 0.0023% in the other cases. As the number of iterations increases, the values in Table 3 converge closer to zero. As described above, the convergence value $< 1\times10^{-6}$ is achieved with 20 repetitions.

Table 2: Comparison of optimal value of $x_1$-$x_5$ (Schwefel function)

| | $x_1$ | $x_2$ | $x_3$ | $x_4$ | $x_5$ |
|---|---|---|---|---|---|
| MOST | 420.959 (-0.002% Error) | 420.959 (-0.002% Error) | 419.983 (-0.23%) | 420.959 (-0.002% Error) | 420.959 (-0.002% Error) |
| Theoretical | 420.969 | 420.969 | 420.969 | 420.969 | 420.969 |

Table 3: Comparison of optimal value of $x_1$-$x_5$ (Sphere function)

| | $x_1$ | $x_2$ | $x_3$ | $x_4$ | $x_5$ |
|---|---|---|---|---|---|
| MOST | $-9.54\times10^{-7}$ | $9.54\times10^{-7}$ | $-9.54\times10^{-7}$ | $-9.54\times10^{-7}$ | $9.54\times10^{-7}$ |
| Theoretical | 0.000 | 0.000 | 0.000 | 0.000 | 0.000 |

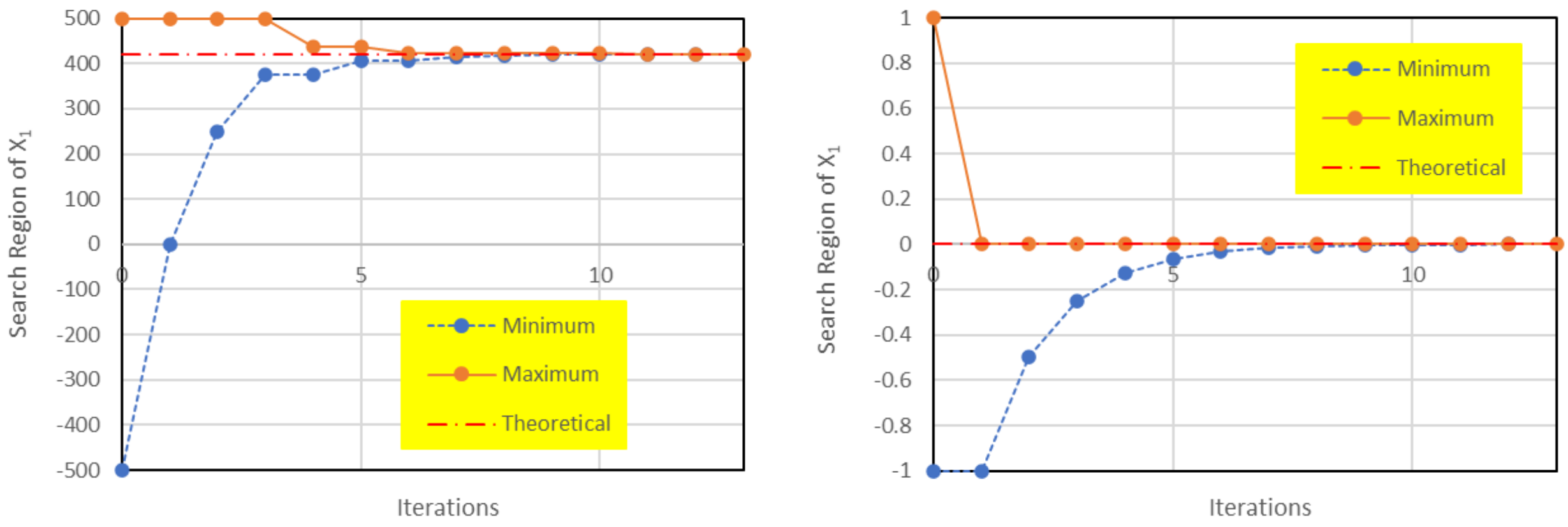

a) Schwefel function　　　b) Sphere function

FIG. 4: Conversion of $X_1$ by MOST

FIG. 4 shows the relationship between the number of iterations and convergence of the variable $x_1$ in Schwefel function and Sphere function. The figure shows that $x_1$ converges toward the theoretical value with the number of iterations in each case. From this, it is considered that the effectiveness of the proposed method: MOST was confirmed in the objective function of the multivariable parameter.

## 3. Application to Neural Network

### 3.1 Neural Network of XOR gate

To represent an XOR gate, consider a neural network with two input layers, two hidden layers, and one output layer shown in FIG. 5 (Mirela Reljan-Delaney and Julie Wall, Fenglei Fan, Wenxiang Cong). A bias 1 is applied to the input layer and the hidden layer, respectively. The XOR gate is a logic circuit in which an output = 1 when ($x_1$, $x_2$) = (0,0), (1,1) is input, and an output = 0 when ($x_1$, $x_2$) = (1,0), (1,0) is output. On a line connecting the nodes, nine weighting factors from $W^1_{11}$ to $W^2_{22}$ are defined as shown in FIG. 5. These coefficients were optimized by Adam and genetic algorithm as existing models for comparison. In addition, the MOST was applied for verification purposes and the results were compared with those of existing models.

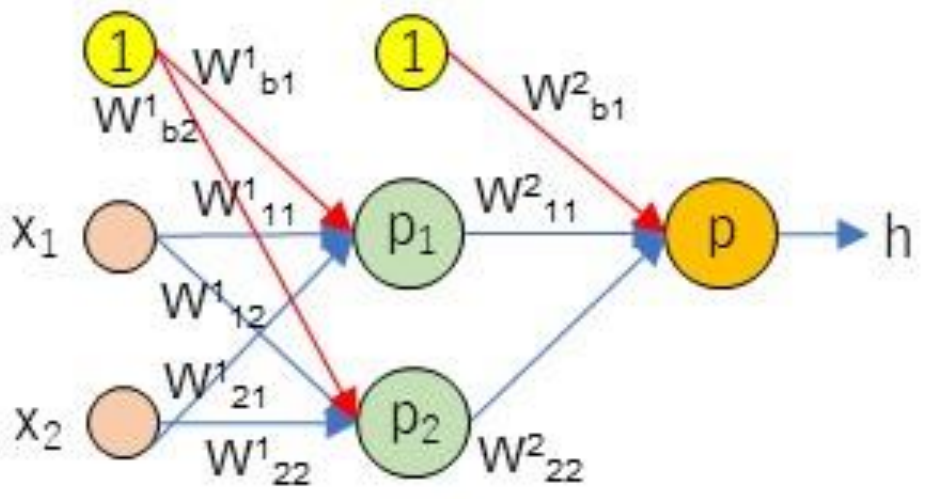

FIG. 5: Configuration of Neural Network of XOR gate

3.2 Benchmarking methods

3.2.1 Adam

In the neural network, in order to update the weight coefficient w to $w^{t+1}$, an error inversion method due to the following gradient method is used.

$$w^{t+1} = w^t - \eta\left(\frac{\partial L}{\partial w}\right)^t \tag{9}$$

Here, η is a hyperparameter, and usually a small value such as 0.001 is given. As a method for accelerating the convergence of w, Adam (Adaptive Moment Estimation) uses the following instead of Eq. (9) in the minimization method (Kingma, D. P., and Ba, J. L.,2015).

$$m^{t+1} = \beta_1 m^t + (1-\beta_1)\left(\frac{\partial L}{\partial w}\right)^t \tag{10}$$

$$v^{t+1} = \beta_2 v^t + (1-\beta_2)\left(\left(\frac{\partial L}{\partial w}\right)^t\right)^2 \tag{11}$$

$$\widehat{m} = \frac{m^{t+1}}{1-\beta_1^{t+1}} \tag{12}$$

$$\hat{v} = \frac{v^{t+1}}{1-\beta_2^{t+1}} \tag{13}$$

$$w^{t+1} = w^t - \eta\frac{\widehat{m}^t}{\sqrt{\hat{v}^t}+\varepsilon} \tag{14}$$

where m (t) is the first moment and ν (t) is the second moment. For other hyperparameters $\eta = 0.001$, $\beta_1 = 0.9$, $\beta_2 = 0.999$, $\varepsilon = 10^{-8}$ are recommended. Adam is said to be efficient in following two ways:

1) The weight factor w is updated at its first moment $m^t$, not at the slope of the loss function ($\partial L/\partial w$). $m^t$ is calculated by Eq. (10), and by averaging ($\partial L/\partial w$) — i.e., removing the noise of ($\partial L/\partial w$), a smooth change in the weighting factor w can be expected.
2) The $\widehat{m}/\left(\sqrt{\hat{v}}+\varepsilon\right)$ in the right side of Eq. (14) tends to have a value close to + 1 or -1 in the early period of learning and close to 0 in the later period. Therefore, it is expected that w is greatly modified at the beginning of learning, and the change of w is small at the end of learning.

In this paper, we compare the results of Adam as a benchmark.

3.2.2 Genetic algorithm (GA)

Genetic algorithm is an algorithm inspired by biological evolution (selective selection, mutation), and can be considered as a technique of probabilistic search (D. Gouldberg, 1989). GA uses three types of genetic operations:

Selection, Crossover, and Mutation. Candidates for solutions are represented one-dimensionally in chromosomes as genotypes. Each generation is a set of individuals. The population size in each generation is called the population size. The processing procedure of GA is as follows.

1. Generation of an initial population
2. Repeat the following until the end condition is met:
   (a) assessment of fitness
   (b) Selection
   (c) Crossover
   (d) Mutation

Genetic algorithms have a long history, and there are various advanced algorithms. This study employs the neighborhood cultivation genetic algorithm (NCGA) (Shinya Watanabe, Tomoyuki Hiroyasu, Mitsunori Miki, 2002). The NCGA has a neighborhood crossover mechanism in addition to the mechanisms of general genetic algorithms. The NCGA is superior to other genetic algorithms in terms of solution surveys in a discrete search space.

3.3 Calculation conditions

Adam used LeRU to suppress gradient loss in the hidden layer, sigmoid function in the output layer, and cross entropy as a loss function. On the other hand, in the GA and the proposed method, a sigmoid function is used for both the hidden layer and the output, and the loss function is an error with the teacher signal. On the Adam, for hyperparameters $\eta$ = 0.001, $\beta_1$ = 0.9, $\beta_2$ = 0.999, $\varepsilon = 10^{-8}$ are applied. The search range of the weight coefficients in the genetic algorithm and this method was -50 to 50 for all coefficients. The genetic algorithm uses 100 individuals and 200 generations. The proposal method used 4000 random numbers for each Monte Carlo method. In the genetic algorithm, an initial value for each coefficient is required and all are given 0.05. In the MOST, it is not necessary to set an initial value, and the search area specification itself of -50 to 50 is the initial value.

Table 4: Specifications of Calculation

| Optimization method | Activation function (Hidden layer) | Activation function (Output layer) | Loss function |
|---|---|---|---|
| Adam | ReLU | Sigmoid function | Cross Entropy |
| Genetic Algorithm | Sigmoid function | | Error=$\frac{1}{2}\sum(Y_i - Y_{t(i)})^2$ |
| Present (MOST) | | | |

3.4 Results of calculations under different optimization

FIG. 6 shows a comparison between the weighting coefficients obtained by the respective optimization methods and the output. This shows that the weighting factors are different for each method. However, since the activation function and loss function used between the methods are different, and even if the same optimization method is used, there are a plurality of different weight coefficients, the difference of the weight coefficients itself is not significant. On the other hand, paying attention to the output in Table 5, in Adam in particular, the output = 0.0032 under the input condition where the output should be 0, which is a larger value than that of the genetic algorithm and this method. On the other hand, in this method, it is approximately $9\times 10^{-9}$, which is sufficiently close to 0.

| | |
|---|---|
| $W^1_{11}$ | 3.27 |
| $W^1_{21}$ | -3.20 |
| $W^1_{b1}$ | -0.09 |
| $W^1_{12}$ | -3.44 |
| $W^1_{22}$ | 3.89 |
| $W^1_{b2}$ | -0.51 |
| $W^2_{11}$ | 4.06 |
| $W^2_{21}$ | 3.84 |
| $W^2_{b1}$ | -5.25 |

a) Adam

| | |
|---|---|
| $W^1_{11}$ | 41.07 |
| $W^1_{21}$ | -49.38 |
| $W^1_{b1}$ | -26.33 |
| $W^1_{12}$ | 49.74 |
| $W^1_{22}$ | -42.98 |
| $W^1_{b2}$ | 15.23 |
| $W^2_{11}$ | 32.17 |
| $W^2_{21}$ | -34.54 |
| $W^2_{b1}$ | 17.39 |

b) Genetic Algorithm

| | |
|---|---|
| $W^1_{11}$ | -49.95 |
| $W^1_{21}$ | 31.20 |
| $W^1_{b1}$ | -21.92 |
| $W^1_{12}$ | -31.20 |
| $W^1_{22}$ | 49.95 |
| $W^1_{b2}$ | 18.02 |
| $W^2_{11}$ | 40.58 |
| $W^2_{21}$ | -37.45 |
| $W^2_{b1}$ | 18.90 |

c) Present (MOST)

FIG. 6 : Comparison of Weight Coefficients

Table 5: Comparison of outputs under different inputs

| Input | Output (Training signal) | Adam | Genetic Algorithm | Present (MOST) |
|---|---|---|---|---|
| $X_1$=0, $X_2$=0 | 0 | 0.0032 | $3.56\times10^{-8}$ | $8.75\times10^{-9}$ |
| $X_1$=1, $X_2$=0 | 1 | 1 | 1 | 1 |
| $X_1$=0, $X_2$=1 | 1 | 1 | 1 | 1 |
| $X_1$=1, $X_2$=1 | 0 | 0.0032 | $3.56\times10^{-8}$ | $8.75\times10^{-9}$ |

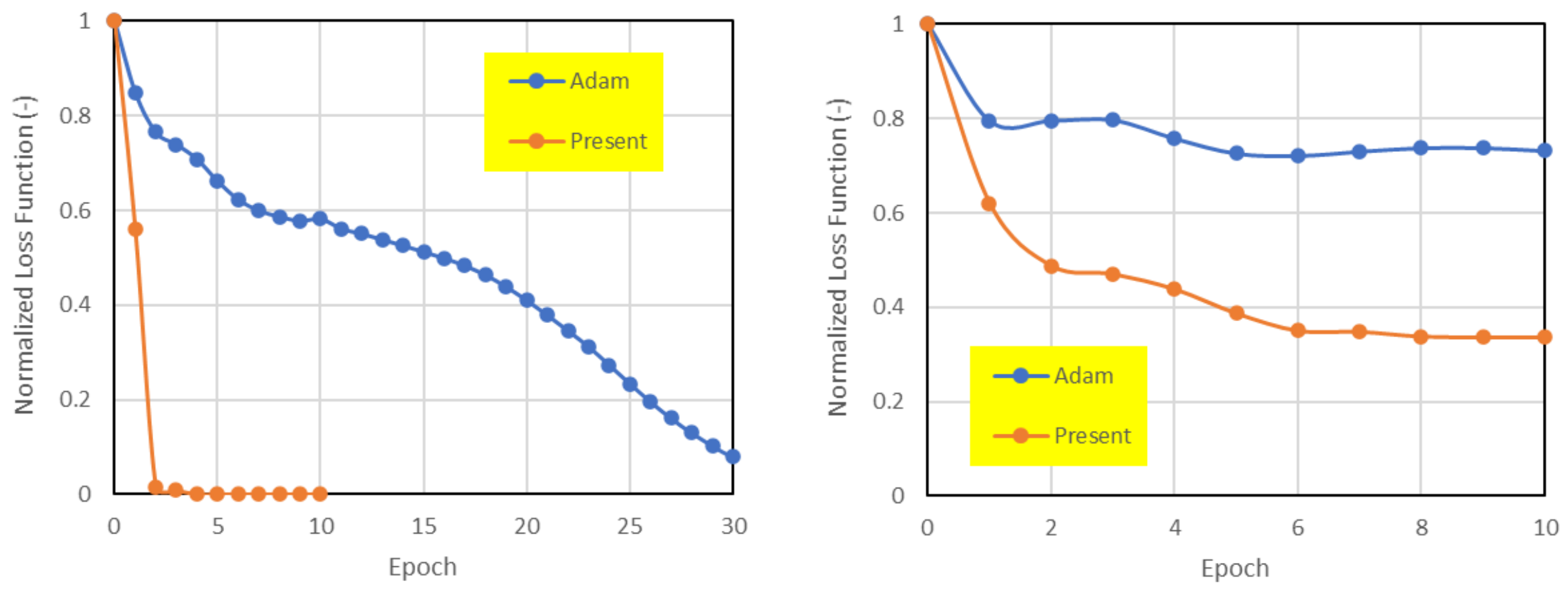

a) Successful learning b) Failed learning

FIG. 7 : Comparison of convergency

A comparison of the convergence between Adam and this method is shown in FIG. 7. FIG. 7a) shows the convergence of the loss function when correctly reproducing XOR gate. Because Adam and the proposed method use different loss functions, they are normalized so that their initial value is 1. This figure shows that the convergence of this method is 3-4 times faster than Adam. From the results of FIGs. 6 and 7a), the MOST was confirmed to be faster and more accurate than Adam. In a neural network, when the number of nodes in a hidden layer is small, a correct optimum solution is often not obtained. FIG. 7b) shows a comparison of convergence in such a case where convergence is not correct. In the case of Adam, all the outputs converged to 0.49 regardless of the input, and in this method, all the outputs converged to 0.63 regardless of the input. In both cases, we can see that the loss function does not asymptotically approach zero and settles to a constant value. This is not an optimization problem, but a neural network specific problem. Therefore, this point cannot be solved by using this method. In the future, the number of nodes in the hidden layer should be increased, and the relationship between the number of nodes and convergence should be evaluated. The calculation time was 10 seconds or less for both Adam (100 Epochs) and this method (10 Epochs), and 3 minutes for GA under the condition of 100 individuals × 300 generations. The convergence of the proposed method is comparable to that of GA. On the other hand, the calculation time was equivalent to that of Adam's case. From this, it is considered that this method can be confirmed to be high-speed and high-precision in application to an XOR gate.

## 4. Conclusions

A new optimization method using the Monte Carlo method was applied to the neural network learning of the XOR gate. As a result, the calculation time is about the same as that of Adam, which is the most applied at present, and the convergence is faster. Comparing the XOR outputs, the MOST results are more accurate than Adam and GA. This shows that a new optimization method using the Monte Carlo method can be applied to neural network learning. In the future, it will be applied to a larger scale neural network learning, and further verification will be attempted. The main features of the MOST are as follows.

1) It is a stochastic method using random numbers, but optimization is a deterministic number without stochastic fluctuation.
2) The solution always converges. In particular, more than 20 MOST iterations converge to $1/10^6$ of the initial search width.
3) It does not include hyperparameters that need to be tuned for optimization. There is also no dimensional mismatch problem to be addressed in the gradient descent method.
4) There is no inherent dimensional mismatch problem in the gradient descent method.

**Appendix:**

1) Knowhow of MOST – Application to an extreme case

Consider the extreme cases in which the objective function changes as shown in FIG. A. If we apply MOST to this case, the integration value is clearly evaluated to be smaller in region [2], in which case the correct solution is not reached. In practice, since the relationship between the objective function and the multivariable parameter is not known, it is not possible to grasp whether the objective function becomes such an extreme before optimization. If such an extreme is expected to occur, it is recommended to apply the two-segment MOST after dividing into 10 regions and selecting 2 -3 regions from them in order to narrow down the initial candidates as a preprocessing.

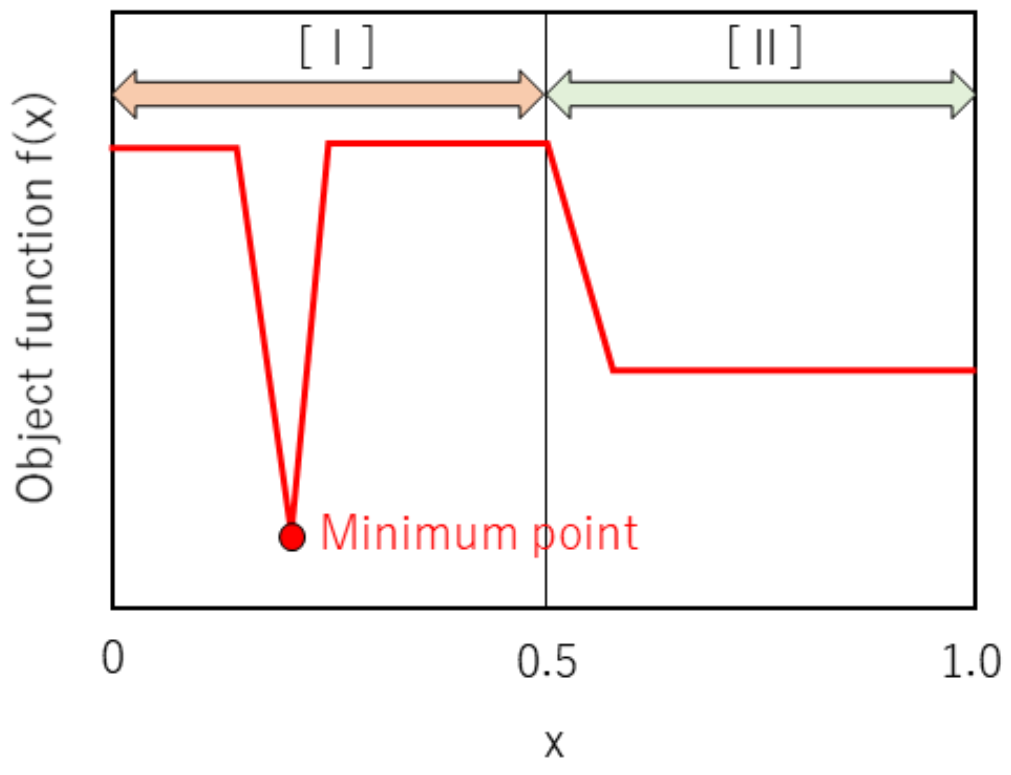

FIG. A.: Extreme case of object function